\documentclass[runningheads]{waica}
\usepackage[T1]{fontenc}
\let\oldthebibliography\thebibliography
\renewcommand{\thebibliography}[1]{
  \oldthebibliography{#1}
  \setlength{\itemsep}{0pt}
  \setlength{\parskip}{0pt}
}

\usepackage{graphicx}
\usepackage{adjustbox}
\usepackage[table,dvipsnames]{xcolor}         

\usepackage[utf8]{inputenc} 
\usepackage[T1]{fontenc}    
\definecolor{iccvblue}{rgb}{0.21,0.49,0.74}
\usepackage[pagebackref,breaklinks,colorlinks,allcolors=iccvblue]{hyperref}
\usepackage{url}            
\usepackage{booktabs}       
\usepackage{amsfonts}       
\usepackage{nicefrac}       
\usepackage{microtype}      

\usepackage{subcaption}

\usepackage{amsmath}
\usepackage{xspace}
\usepackage{adjustbox}
\usepackage{multirow}
\usepackage{bbding}
\usepackage{booktabs}       
\usepackage[normalem]{ulem}
\usepackage{graphicx}

\usepackage{array}
\usepackage{relsize}
\usepackage[capitalize]{cleveref}
\usepackage{tikz}

\newcommand{\system}{\textsc{M-CPT}\xspace}
\usepackage{arydshln}

\crefname{section}{Sec.}{Secs.}
\Crefname{section}{Section}{Sections}
\Crefname{table}{Table}{Tables}
\crefname{table}{Tab.}{Tabs.}

\newcommand{\tablestyle}[2]{\setlength{\tabcolsep}{#1}\renewcommand{\arraystretch}{#2}\centering}

\usepackage{makecell}
\newcommand{\MissedData}{-\xspace}

\newcommand{\ourpe}{\textsc{CPE}\xspace}

\def\eg{\emph{e.g.}\xspace} 

\def\ie{\emph{i.e.}\xspace}

\begin{document}
\title{Enhancing Vision Foundation Models via Multimodal Continual Pre-Training}
\titlerunning{\system}


\author{
Yitong Chen\inst{1,2}\textsuperscript{*}\and
Lingchen Meng\inst{1}\textsuperscript{*}\and
Wujian Peng\inst{1,2}\and
Jun Tao\inst{3}\and
Chenjie Xu\inst{3}\and
Zuxuan Wu\inst{1,2}\textsuperscript{\(\dagger\)}\and
Yu-Gang Jiang\inst{1}
}

\authorrunning{Chen et al.}

\institute{Institute of Trustworthy Embodied AI, Fudan University\and 
Shanghai Innovation Institute\and Shanghai Baosight Software Co., Ltd.}

\maketitle              

\begingroup
\renewcommand{\thefootnote}{}
\footnotetext{\textsuperscript{*}Equal contributions.}
\footnotetext{\textsuperscript{\(\dagger\)}Corresponding author.}
\endgroup

\begin{abstract}
Vision Foundation Models (VFMs) provide strong visual representations for a wide range of applications. 
In this work, we enhance prevailing VFMs through multimodal training, allowing them to effectively process visual inputs at varying resolutions while producing visual representations that are better aligned with language representations, regardless of their original pre-training objectives. 
To this end, we introduce \system, a \textbf{M}ultimodal \textbf{C}ontinual \textbf{P}re-\textbf{T}raining framework designed to improve the understanding capability of pretrained VFMs while preserving their strong visual representation quality. \system introduces a Continual Position Embedding (CPE) for handling flexible visual resolutions, along with a feature alignment objective that improves the consistency between visual and textual representations during multimodal training.
Extensive experiments on leading VFMs, including DINOv2, SigLIP, and AIMv2, demonstrate that \system consistently improves multimodal understanding performance while preserving strong performance on standard vision benchmarks such as classification and segmentation. The code is available \href{https://github.com/ShareLab-SII/CoMP-MM}{here}.



\end{abstract}    
\section{Introduction}
\label{sec:intro}

Pre-training Vision Foundation Models (VFMs) capable of extracting transferable representations for various downstream tasks has been a long pursuit of the computer vision community. The key to pre-training is to scale up models and data through constructing strong supervisory signals with weak-strong augmentations in vision-only pretraining~\cite{dino,dinov2,he2020moco,chen2020simclr} or cross-modality alignment in vision-language pre-training~\cite{clip,siglip,siglip2,aimv2}. These VFMs often demonstrate strong performance for a variety of downstream tasks, and can be combined with Large Language Models (LLMs) by designing adapters that project visual features into text space.




In this paper, we revisit these widely used VFMs such as vision-only pre-training DINOv2~\cite{dinov2}, vision-language pre-training SigLIP~\cite{siglip} and AIMv2~\cite{aimv2}. We argue that these prevailing VFMs, regardless of their pre-training procedures, can be further boosted through continual multimodal pre-training. This allows VFMs to (1)  better process visual inputs of arbitrary sizes without requiring resizing, when used as vision encoders of LMMs; (2) produce outputs that are more aligned with language representations, thereby improving multimodal understanding,  significantly benefiting encoders from vision-only pre-training. 

\begin{figure*}[t!]
    \centering
    \includegraphics[width=0.9\linewidth]{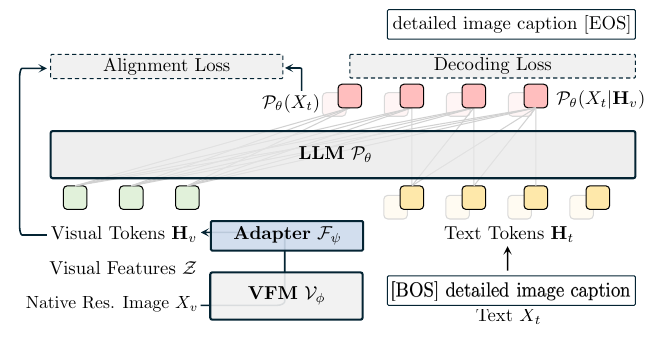}
    \vspace{-1mm}
    \caption{\small
    \textbf{Overview of \system.} Our method accepts an image at native resolution and its corresponding text. Then, in addition to training through text decoding in next-token prediction paradigm, we also explicitly project the visual features into the language space of LLM using Alignment Loss.} 
    \label{fig:arch}
\end{figure*}

On the one hand, equipping VFMs with the ability to deal with images of varying sizes is the core of visual understanding, as image resolutions directly impact the richness of the information within an image. However, existing approaches treat ``an image as worth 16×16 words"\cite{vit}, resizing all images to a predefined size. This one-size-fits-all strategy results in the loss of critical details, impeding the model’s ability to perceive fine-grained information. This is particularly detrimental in tasks that demand high-resolution inputs, such as chart understanding\cite{chartqa}, document parsing~\cite{docvqa}, and fine-grained recognition~\cite{instit}. 
Recent works~\cite{flexivit,resformer,dinov2,siglip,siglip2,aimv2} have attempted to address this challenge by improving bilinear interpolation of positional embeddings and incorporating multi-resolution training. However, they still struggle with real-world scenarios involving diverse input resolutions, as they remain constrained by the limitations of extrapolation.


On the other hand, we believe there is still a representation gap between VFMs and LLMs, which results from their distinct training objectives and data modalities during pre-training~\cite{mindthegap}. To bridge this gap and enable LMMs to better understand visual inputs, the mainstream approach~\cite{llava,llavanext,llavaov} involves training an adapter that projects the visual embeddings of VFMs into the textual embedding space of LLMs, typically through next-token prediction for text tokens. However, relying solely on text-based supervision is insufficient to effectively and directly reduce this gap~\cite{vlap}, particularly when VFMs have not undergone vision-language alignment pretraining~\cite{cambrain,zong2024mova}.

To address these challenges, we introduce \system, a continual pre-training pipeline that is carefully designed to enhance exisiting VFMs. Specifically, \system builds upon (1) \ourpe, a Continual Position Embedding for vision models, which is operated by adding the standard RoPE-2D~\cite{rope2d} with the learned 1D position embedding, to support native resolution continual pre-training; (2) Alignment Loss, a cross-entropy loss between visual and textual features through language prototypes, to align representations between VFMs and LMMs. 

As shown in \cref{fig:arch}, our method accepts an image with native resolution and its corresponding text. Then, in addition to training through next-token prediction on the text, we also explicitly project the visual features into the language space using Alignment Loss by word embedding of LLM. With a three-stage continual pre-training, our models excel not only on multimodal understanding, but also in downstream tasks such as classification and segmentation. 


\section{\system}
\label{sec:method}
\vspace{-1mm}
Our goal is to empower a pre-trained vision foundation model (VFM) with the ability to process images at their native resolution while aligning its encoder features with the representation space of a pre-trained LLM. To achieve this, as shown in \cref{fig:arch}, we propose a continual multimodal pre-training pipeline that improves existing VFMs so that they can handle native resolution inputs using \ourpe~(\cref{method:native}) and better align with language space through carefully designed loss functions: the text decoding loss~(\cref{method:decode}) and the cross-modal alignment loss~(\cref{method:align}).
These components are integrated into a three-stage training (\cref{method:recipe}), ensuring effective adaptation and alignment.



\subsection{Native resolution adaptation}
\label{method:native}
Vision encoders often use fixed-size inputs during the pre-training stage, and thus struggle to handle images with varying resolutions, particularly high-resolution images for fine-grained visual understanding. While training on images of different sizes is a straightforward solution, it is particularly challenging  due to the predefined shape of position embeddings in vision transformers. A common approach is to interpolate the original position embeddings in an online manner to accommodate different input resolutions, yet the results are unsatisfactory~\cite{flexivit,resformer}.

Inspired by the success of Rotary Position Embedding (RoPE)~\cite{rope} that demonstrates strong extrapolation capabilities in NLP~\cite{Touvron2023LLaMAOA,xiong2024effective} and CV~\cite{navit,pixtral,qwen2vl,qwen2.5vl,ropevision}, we aim to build upon RoPE-2D~\cite{rope2d} to handle a sequence of visual tokens. Unlike previous methods that only rely on RoPE-2D, which is neither a data-efficient nor a training-friendly approach (see \cref{tab:ab_pe}), we leverage both absolute and relative embeddings to ensure a smooth transition from the pre-trained vanilla ViT to arbitrary resolutions, and capture richer positional information so as to handle a variety of high-resolution inputs. We refer to this combination as \ourpe.



\begin{figure}[ht!]
    \centering
    \includegraphics[width=1.0\linewidth]{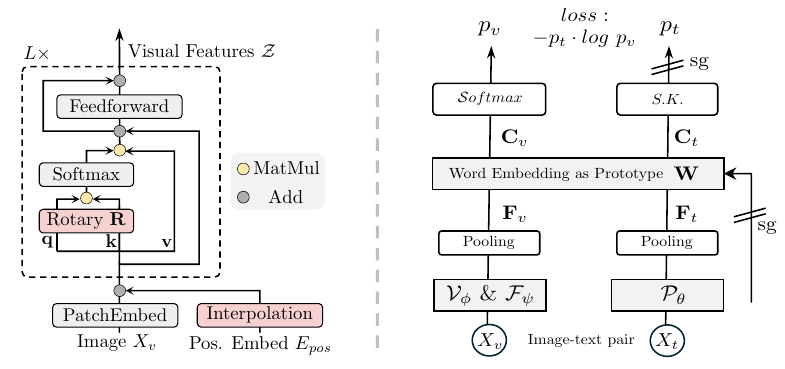}
    \caption{\small \textbf{Left: \ourpe}. The projection layers $\mathcal{P}roj_{q,k,v,o}$ and scale operators are omitted. We leverage both absolute learned position embedding and relative RoPE-2D~\cite{rope2d} to capture richer positional information. \textbf{Right: Alignment Loss}. We illustrate it in the case of one single pair of global vision and text features $\mathbf{F}_v$ and $\mathbf{F}_t$ for simplicity. $\mathbf{F}_v$ and $\mathbf{F}_t$ are mapped by frozen learned prototype $\mathbf{W}$, \ie, the word embedding of LLMs. Then, they are converted into normalized probabilities using the Softmax function and iterative Sinkhorn-Knopp algorithm~\cite{sk}. Finally, cross-entropy is applied as the loss. To prevent information leakage, the text features are extracted without image prefixes.}
    \vspace{-4mm}
    \label{fig:method}
\end{figure}

Specifically, as shown in \cref{fig:method} (left), a 2D image of resolution $(H, W)$ is patchified into $N=HW/P^2$ patches $x_p \in \mathbb{R}^{N\times (P^2\cdot C)}$, where $P$ denotes the patch size of vision encoder and $C$ indicates the number of image channel. The image encoding processes can be expressed as:
\begin{align}
    &\mathbf{z}_0 = [\mathbf{x}_p^1\mathbf{E};\mathbf{x}_p^2\mathbf{E};\cdots;\mathbf{x}_p^N\mathbf{E}] + \mathcal{I}nt(\mathbf{E}_{pos}),\\
    &\mathbf{q}_i, \mathbf{k}_i, \mathbf{v}_i = \mathcal{P}roj_q(\mathbf{z}_i), \mathcal{P}roj_k(\mathbf{z}_i), \mathcal{P}roj_v(\mathbf{z}_i), \\
    &\mathbf{y}_i = \mathbf{z}_i + \mathcal{P}roj_{o}(\mathcal{S}oftmax(\frac{(\mathbf{Rq}_i)^\top(\mathbf{Rk}_i)}{c})\mathbf{v}) \label{eq:rope},\\
    &\mathbf{z}_{i+1} = \mathbf{y}_i + \mathcal{FFN}(\mathbf{y}_i), \quad where \ i = 0,\cdots ,L-1 \label{eq:feat},
\end{align}
where $E \in \mathbb{R}^{(P^2\cdot C \times D_v)}$ and $E_{pos} \in \mathbb{R}^{N\times D_v}$ indicate patch embedding and learnable position embedding, $D_v$ indicates the hidden dimension of vision encoder, $\mathcal{I}nt(\cdot)$ represents bilinear interpolation, $\mathcal{P}roj(\cdot)$ is the projection layer, $\mathcal{FFN}(\cdot)$ denotes standard feed-forward network, and $L$ denotes the number of encoder layers. In particular, $\mathbf{R}$ in \cref{eq:rope} is the 2D rotary matrix\cite{rope2d}:
\[
\mathbf{R}_{x,y} = \left(\begin{array}{cc:cc}
    \cos x\theta & -\sin x\theta & 0 & 0 \\
    \sin x\theta & \cos x\theta & 0 & 0 \\
    \hdashline
    0 & 0 & \cos y\theta & -\sin y\theta \\
    0 & 0 & \sin y\theta & \cos y\theta
\end{array} \right).
\]

\subsection{Text-supervised generative pre-training}
\label{method:decode}

Text-supervised generative pre-training is widely used in large multimodal models (LMMs). It extends the standard text-only autoregressive next-token prediction framework~\cite{radford2018improving,radford2019language,Brown2020LanguageMA} to visual inputs, mapping visual features into the input layer of a large language model via query-driven cross-attention~\cite{Li2023BLIP2BL,dai2024instructblip,alayrac2022flamingo} or projection~\cite{llava}. We adopt the projection-based multimodal framework due to its simplicity and effectiveness. Formally, the projection $\mathcal{F_\psi}$ can be defined as:
\begin{equation}
    \mathbf{H}_v = \mathcal{F_\psi}(\mathcal{Z}),
\end{equation}
where $\mathcal{Z} = \mathbf{z}_L$ in \cref{eq:feat}. Then, the image tokens $\mathbf{H}_v$ are fed into the input layer of the LLM $\mathcal{P_\theta}$, serving as the condition for the corresponding text $X_t$. The text decoding loss can be expressed as:
\begin{equation}
    L_{dec} = -\frac{1}{T}\sum_{i=V+1}^{V+T}\log \mathcal{P_\theta}(X_i|X_{<i},\mathbf{H}_v),
\end{equation}
where $V$ and $T$ denote the number of visual and textual tokens, respectively. To support the autoregressive generation, the image-grounded text decoder utilizes causal self-attention mechanisms.

\subsection{Vision-language representation alignment}
\label{method:align}
Thanks to text-supervised generative pre-training and decoding loss, the vision encoder can be optimized using paired images and texts. However, the supervision is too distant for directly optimizing the vision model, especially when the original pre-training of vision encoder does not involve vision-language alignment, such as in image-only SSL of DINOv2~\cite{dinov2} (see \cref{tab:ab_recipe}). To bridge the modality gap between the vision and language models, we utilize direct vision-language representation alignment by encoding visual and textual features through the VFMs and LLMs.

Some previous approaches~\cite{internvl,Yu2022CoCaCC,lcl} employ a contrastive loss for cross-modal alignment, which usually requires large batch sizes and even an auxiliary text encoder. To avoid these constraints and inspired by the knowledge distillation in DINO~\cite{dino,dinov2}, we align visual and textual representations by treating the word embeddings $\mathbf{W}$ of the LLM as prototypes~\cite{vlap}. Specifically, as shown in \cref{fig:method} (right), we first obtain global visual and its corresponding text features $\mathbf{F}_v$ and $\mathbf{F}_t \in \mathbb{R}^{B \times D_t}$ through the VFM and LLM by parameter-free global average pooling, respectively, as follows:
\begin{equation}
    \mathbf{F}_v=Pool(\mathbf{H}_v), \mathbf{F}_t=Pool(\mathcal{P_\theta}(X_t)),
\end{equation}
where $B$ denotes the mini-batch size, $D_t$ represents the hidden dimension of LLM and $X_t$ is a piece of corresponding text to $\mathbf{H}_v$. And then, we map $\mathbf{F_v}$, $\mathbf{F_t}$ into language space by the prototype $\mathbf{W} \in  \mathbb{R}^{D_t\times K}$ as follows:
\begin{equation}
    \mathbf{C}_v=\mathbf{W^\top F}_v, \mathbf{C}_t=\mathbf{W^\top F}_t,
\end{equation}
where $K$ indicates the vocabulary size of LLM. To prevent information leakage during training, the text features are extracted without image prefixes. Additionally, we detach the gradient of the word embedding to avoid training collapse.

Moreover, for adapting the learned prototype, we replace the Softmax function in DINO, which assumes a uniform distribution, with iterative Sinkhorn-Knopp algorithm~\cite{sk} that explores the prior distribution of word embeddings~\cite{vlap} to obtain soft normalized probabilities of $\mathbf{C}_t$:
\begin{equation}
    p_t = \mathcal{D}iag(\mathbf{u}_W)exp(\frac{\mathbf{C}_t}{\epsilon}) \mathcal{D}iag(\mathbf{v}),
    \label{eq:algloss}
\end{equation}
where $\mathbf{u}_W \in \mathbb{R}^{K} $ is the prior marginal distribution of words, and $\mathbf{v}\in \mathbb{R}^{B} $ are renormalization vectors. Thus, the alignment loss can be formally expressed as:
\begin{equation}
    L_{align} = -p_t\cdot \log\ p_v,
\end{equation}
where, $p_v=\mathcal{S}oftmax(\mathbf{C}_v)$. Additionally, we stop the gradient propagation for LLM, ensuring that $L_{align}$ updates only the parameters of VFM.

\begin{table*}[!t]
    \footnotesize
      \centering
      \tablestyle{0.1pt}{1.0}
      \caption{ \small \textbf{Main results of \system-MM on multimodal understanding benchmarks.} \#PT indicates the size of pre-training dataset. \#IT indicates the size of intrcution tuning dataset. N/A indicates the size is unknown. $\dagger$ denotes we report the performance on validation sets. $\star$ denotes we replace original SigLIP with our \system-SigLIP for initialization.}
      \begin{adjustbox}{width=\linewidth}
      \begin{tabular}{l cccc ccc c  c c}
      \toprule
      \multirow{2}{*}{Model} & \multirow{2}{*}{\#PT} & \multirow{2}{*}{\#IT \quad}  & \multicolumn{4}{c}{\textcolor{gray}{\textit{Text-rich and Fine-grained}}} &\multicolumn{4}{c}{\textit{\textcolor{gray}{General and Real-world}}}  \\
      &&& CQA & DVQA$^\dagger$ & AI2D  & INST & \quad   GQA & MMMU$^\dagger$ & MMB$^\dagger$ & RWQA \\
      \midrule
       \multicolumn{11}{c}{\textit{$\sim$1B models}} \\ 
      
      Deepseek-VL~\cite{deepseekvl} & 3.75M & N/A  & - & - & 51.5 & - & -  &32.2 & - & - \\
      
      LLaVA-OV (SI)~\cite{llavaov} & 4.6M & 3.2M  & 61.0 &  {75.0} & 54.2 & 44.2 & -&31.2 & 43.8 & 53.7 \\
      LLaVA-OV~\cite{llavaov} & 4.6M & 4.8M  &  {61.4} & 73.7 &  {57.1} &  {47.8} &- & {31.4} &  {52.1} &  {55.6}  \\
      \rowcolor{RoyalBlue!3}
      \system-MM-DINOv2 & 4.6M & 3.2M  & 66.1 & 71.0 & 60.0 & 48.7 & 59.3  & 30.4 & 44.7 & 57.3  \\
      \rowcolor{RoyalBlue!3}
      \system-MM-SigLIP & 4.6M & 3.2M  & 66.7 & 75.9 & 61.9 & 50.1 &  60.4 & 33.0 & 56.4 & 58.3  \\
      \rowcolor{RoyalBlue!3}
      \system-MM-AIMv2 & 4.6M & 3.2M  &  68.3 & 75.8 & 64.9 & 50.3 & 61.6 & 35.7 & 61.6 & 58.7 \\
      \midrule
        \multicolumn{11}{c}{\textit{$\sim$7B models}} \\ 
      LLaVA-1.5~\cite{Liu2023ImprovedBW} & 558K & 665K &18.2 & 28.1 & - & 32.1  & 62.0 & 35.3 & - & - \\
      LLaVA-NeXT~\cite{llavanext} & 558K & 765K & 54.8 & 74.4 & - & 42.4  & 64.2 & 35.1 & - & - \\
      Deepseek-VL~\cite{deepseekvl} & 3.75M & N/A  & - & - & 65.3 & - &  - & 36.6 & - & - \\
      Cambrian-1~\cite{cambrain} & 1.2M & 7.0M  & 73.3 & 77.8 & 73.0 & -  & 64.6 & 42.7 & - & 64.2 \\
      LLaVA-OV (SI)~\cite{llavaov} & 4.6M & 3.2M  & 78.8 & 89.3 & 81.6 & 61.8 & - &47.3 & 81.7 & 65.5  \\
      LLaVA-OV~\cite{llavaov} & 4.6M & 4.8M  & 80.0 & 90.2 & 81.4 & 71.7 &-&  {48.8} & 80.8 & 66.3  \\
      \rowcolor{RoyalBlue!3}
      \system-MM-SigLIP & 4.6M & 3.2M  &  79.6 & 91.0 &  81.4 &  65.0 & 65.9& 48.9 &  81.4 & 66.4  \\
      \rowcolor{RoyalBlue!3}
      \system-MM-SigLIP$^\star$ & 4.6M & 3.2M  &  81.8 & 92.3 & 81.7 & 68.2 & 65.8 & 47.8 & 81.4 & 66.2 \\
      
      \bottomrule
      \end{tabular}
      \end{adjustbox}
    \label{tab:main_lmm}
\end{table*}

\subsection{Training recipe}
\label{method:recipe}
Our continual pre-training is divided into three stages:

\noindent\textbf{Stage-I: Vision-language adapter warming up.} We freeze the VFM and the LLM, and only train the adapter at fixed low image resolution without RoPE-2D.

\noindent\textbf{Stage-II: Native resolution adaptation.} We train the entire model using RoPE-2D at a fixed high resolution, followed by a period at native resolution.

\noindent\textbf{Stage-III: Instruction tuning (optional).} We fine-tune the entire model on the instruction dataset at native resolution with RoPE-2D.






\section{Empirical results}
\label{sec:exp}

We evaluated the performance of our \system-MM on multimodal benchmarks with other LMMs. Additionally, we conducted detailed comparisons with other VFMs across various visual downstream tasks, including multimodal understanding, image classification, and semantic segmentation.

\begin{table*}[!t]
      \centering
      \tablestyle{0.6pt}{1.1}
      \caption{ \small \textbf{Evaluation on multimodal understanding benchmarks.} We conduct extenseive experiments on \system-DINOv2-Large, \system-SigLIP-So400M and \system-AIMv2-Large with LLaMA-3.0 8B~\cite{llama3}, freezing the vision encoder and directly tuning on LLaVA SFT data~\cite{llava} for one epoch. \#Patch indicates the number of input visual patches for the LLM. $\dagger$ denotes we report the performance on validation sets. We also report the performance of the SigLIP 2~\cite{siglip2} NaFlex variant. }
      \begin{tabular}{lc cccccccc}
      \toprule
      Model & ViT & \#Patch  & OKVQA$^\dagger$ & TVQA$^\dagger$ & DVQA & IVQA & CQA & SEED & MME  \\
      \midrule

      DINOv2~\cite{dinov2} & L/14 & 576 & 54.1 & 13.4 & 7.3 & 21.3 & 10.8 & 57.0 & 1345  \\
      DINOv2~\cite{dinov2} & G/14 & 3034  & 56.9 & 15.1 & 8.2 & 19.7 & 12.0 & 68.9 & 1423 \\
    \rowcolor{RoyalBlue!3}
      \system-DINOv2 & L/14 & 576  & 59.0 & 53.6 & 24.7 &22.8 &  {23.8} & {72.8} &1484 \\
      \midrule
      CLIP~\cite{clip} & L/14 & 576  & 60.0 & 47.5 & 25.6 & 21.8 & 19.2 & 70.1 & 1481 \\
      SigLIP~\cite{siglip} & L/14 & 576  & 59.3 & 44.1 & 16.9 & 20.7 & 14.4 & 66.8 & 1416 \\
        SigLIP~\cite{siglip} & So/14 & 729  & 60.1 & 47.5 & 19.2 & 21.0 & 14.7 & 67.5 & 1433 \\
      SigLIP 2 (NaFlex)~\cite{siglip2} & So/16 & 576  & 60.6 & 59.9 & 28.9 & 25.0  & 18.4 & 73.1 & 1536 \\
      \rowcolor{RoyalBlue!3}
      \system-SigLIP & So/14 & 576 &  {61.0} & 62.5 & 34.0 & 26.0 & 25.0 & 74.3 &  {1543} \\
      \midrule
      AIMv2 (336px)~\cite{aimv2} & L/14 & 576  & 60.8 & 53.6 & 26.6 & 22.8 & 19.2 & 71.8 & 1472 \\
      AIMv2 (336px)~\cite{aimv2} & H/14 & 576  & 61.3 &  {55.5} &  {27.8} &  {23.1} & 19.9 & 72.1 & 1545 \\
      \rowcolor{RoyalBlue!3}
      \system-AIMv2 & L/14 & 576 & 60.9 & 60.7 & 33.4 & 24.9 & 23.6 & 73.2 & 1520\\
      \bottomrule
      \end{tabular}
    \label{tab:main_lmm_eval}
\end{table*}

\subsection{\system-MM}

\noindent\textbf{Setup.}
For 1B models, we utilize DINOv2-L~\cite{dinov2}, SigLIP-So400M~\cite{siglip}, AIMv2-L-336px~\cite{aimv2} as pre-trained VFMs and Qwen2.5-0.5B~\cite{qwen2.5} as pre-trained LLM; for 7B models, we utilize original SigLIP-So400M and \system-SigLIP-So400M obtained by 1B model as VFM, and Qwen2.5-7B~\cite{qwen2.5} as LLM. The cross-modality adapter is a 2x2 downsampling MLP. For Stage-I, we train the adapter on LLaVA-Pretrain data~\cite{llava} at pre-trained visual resolution. For Stage-II, we train the full model on LLaVA-Mid-Stage data~\cite{llavaov} at 1024px and native resolution. To support high resolution inputs, we replace 1M data in CC3M~\cite{cc3m} with Densefusion-1M~\cite{densefusion}. For Stage-III, we train the full model on LLaVA-OV-SI SFT data~\cite{llavaov} at native resolution. All experiments are conducted on 8 $\times$ H100.

\noindent\textbf{Results.}
As shown in \cref{tab:main_lmm}, under the slimilar pre-training data size, our model significantly outperforms all other methods and achieves state-of-the-art performance among \textbf{open-data-source} models across multiple benchmarks and on both 1B models and 7B models. Specifically, \system outperforms LLaVA-OV-SI~\cite{llavaov}, a strong baseline that employs the AnyRes technique for high-resolution input, not only on text-rich and fine-grained understanding tasks such as ChartQA~\cite{chartqa}, DocVQA~\cite{docvqa}, AI2D~\cite{ai2d}, and Inst-IT~\cite{instit}, but also on various general and real-world multimodal understanding tasks like GQA~\cite{hudson2019gqa}, MMMU~\cite{mmmu}, MMBench~\cite{mmbench} and RealWorldQA~\cite{realworldqa}.

\subsection{Multimodal understanding}

\noindent\textbf{Setup.}
To further quantify the performance of \system for LMMs, we compare it with other mainstream VFMs following the settings and hyperparameters in~\cite{aimv2}. Specifically, we reinitialize an adapter between \system vision encoder and LLM, \eg LLaMA 3.0 8B~\cite{llama3}, and freeze the parameters of vision encoder all the time. We train the adapter and the LLM jointly in a single stage on LLaVA SFT data~\cite{llava} for one epoch, and scale up the learning rate of adapter by a factor of 8. To ensure fairness, we used the checkpoint before instructing tuning (Stage III) to confirm that the model had not been exposed to instruction tuning data and fixed the number of patches input to the LLM at 576.

\noindent\textbf{Results.}
We evaluate \system across various benchmarks covering general knowledge (OKVQA~\cite{schwenk2022okvqa}, SEED-Bench~\cite{li2023seed}, MME~\cite{mme}) and text-rich (TextVQA~\cite{singh2019towards}, DocVQA~\cite{docvqa}, InfoVQA~\cite{mathew2022infographicvqa}, ChartVQA~\cite{chartqa}) tasks. As presented in \cref{tab:main_lmm_eval}, our models outperform DINOv2, SigLIP and AIMv2 by a significant margin. 

\begin{table}[t]
  \centering
  \caption{ \small \textbf{Results on visual downstream tasks.}}
  \begin{subtable}[t]{0.5\linewidth}
    \tablestyle{2pt}{1.033}
        \caption{ \small \textbf{Evaluation on frozen trunk classification.} All experiments are conducted on ImageNet-1K~\cite{imagenet} at 224px and 448px by utilizing attentive pooling probing.}
      \begin{tabular}{l  c  c  c  }
      \toprule
      Model & ViT  &224px & 448px \\

      \midrule
      MAE~\cite{mae} & H/14 & 78.5 & \MissedData \\
      DINOv2~\cite{dinov2} & L/14  & 86.3 & 87.6 \\
      \rowcolor{RoyalBlue!3}
      \system-DINOv2 & L/14  & 85.7 & 86.5 \\
      \midrule
      CLIP~\cite{clip} & L/14  & 84.4 & 83.8 \\
      SigLIP~\cite{siglip} & So/14  & 87.1 & 88.2  \\
      LLaVA-SigLIP~\cite{llavaov} & So/14 & 83.2 & 84.4 \\
      \rowcolor{RoyalBlue!3}
      \system-SigLIP & So/14  & 86.5 & 87.4 \\
    \midrule
      AIMv2 (224px)~\cite{aimv2} & L/14  & 86.6 & 84.8  \\
      AIMv2 (448px)~\cite{aimv2} & L/14  & 78.9 & 87.9  \\
      \rowcolor{RoyalBlue!3}
      \system-AIMv2 & L/14  & 86.1 & 87.3 \\
      \bottomrule
      \end{tabular}

    \label{tab:main_cls}
  \end{subtable}
  \hfill
  \begin{subtable}[t]{0.47\linewidth}
    \centering
      \tablestyle{0.5pt}{1.15}
      \caption{ \small \textbf{Evaluation on semantic segmentation.} All experiments are conducted on ADE20K~\cite{ade20k} at 504px and 672px by freezing the backbone and only train the UperNet~\cite{upernet} head.}
      \begin{tabular}{l c c  c }
      \toprule
      Model & ViT& 504px & 672px \\
      \midrule
      DINOv2~\cite{dinov2} & L/14 &55.3 & 55.9 \\
      \rowcolor{RoyalBlue!3}      
      \system-DINOv2 & L/14 & 52.7 & 53.0 \\
      \midrule
      SigLIP~\cite{siglip} & So/14 & 35.2 & 31.6 \\
      SigLIP 2 (NaFlex)~\cite{siglip2} & So/16 &  35.3 & 34.8 \\
      LLaVA-SigLIP~\cite{llavaov} & So/14 &  39.9 & 36.5 \\
      \rowcolor{RoyalBlue!3}
      \system-SigLIP & So/14 & 49.5 & 49.1 \\
       \midrule
      AIMv2 (336px)~\cite{aimv2} & L/14  & 51.5 & 50.2  \\
      \rowcolor{RoyalBlue!3}
      \system-AIMv2 & L/14  & 51.0 & 51.8 \\
      \bottomrule
      \end{tabular}

    \label{tab:main_seg}
    
  \end{subtable}
\end{table}

\vspace{-4mm}
\subsection{Image recognition}
\vspace{-1mm}

\noindent\textbf{Setup.}
We evaluate the global-view semantic quality of \system by image classification on ImageNet-1K~\cite{imagenet}. In detail, we utilize attentive pooling probing, \ie adding an attention pooling layer on top of the frozen features, to train our method at fixed 224px and 448px. To further analyze the recognition performance of our model, we evaluate a SigLIP variant from the checkpoint of LLaVA-OV-SI-0.5B model, denoting it as LLaVA-SigLIP. All our vision models are from Stage-II of \system-MM-1B.

\noindent\textbf{Results.}
As shown in \ref{tab:main_cls}, our model preserves the rich global features of the original VFMs while supporting native resolution inputs. Notably, our \system-SigLIP also outperforms LLaVA-SigLIP from LLaVA-OV-SI-0.5B checkpoint by a large margin, indicating our approach can better preserve the classification ability after the continual pre-training.

\begin{table}[!t]
      \centering
      \tablestyle{2pt}{1.0}
      \caption{ \small \textbf{Ablation on training recipe.} The first row presents the baseline three-stage training recipe, with progressive resolutions of 384px, 576px, and 768px. In the first stage, only the adapter is unfrozen, while in the subsequent stages, the full model is fine-tuned. Each following row modifies strategy from the last non-italicized row, and our final recipe is highlighted in \colorbox{RoyalBlue!10}{blue}. Experiments are run using SigLIP-So400M and DINOv2-Large with Qwen2.5-0.5B. $\dagger$: Unfreezing vision encoder for RoPE; $\Delta$: The average of the difference from  previous row.}
      \begin{tabular}{l l   c  c  c  c}
      \toprule
      \# & & AI2D & ChartQA &  DocVQA & $\Delta$ \\
      \midrule
      1 & Baseline Recipe & 50.2 & 29.9 & 24.1 & - \\
      \rowcolor{gray!2}
      \textcolor{gray}
      2 & \hspace{1.5mm} \textcolor{gray}{\textit{+ RoPE-2D from Stage-I${\dagger}$}} & \textcolor{gray}{\textit{51.5}} & \textcolor{gray}{\textit{11.0}} & \textcolor{gray}{\textit{11.4}} & \textcolor{gray}{\textit{$\downarrow$ 10.1}}\\
      3 & \hspace{1.5mm} + RoPE-2D from Stage-II & 55.4 & 56.9 & 61.6 & \textcolor{ForestGreen}{$\uparrow$ 23.2}\\
      \rowcolor{gray!2}
      \textcolor{gray}
      4 & \hspace{1.5mm} \textcolor{gray}{\textit{+ Freezing LM in Stage-II}} & \textcolor{gray}{\textit{54.6}} & \textcolor{gray}{\textit{55.4}} & \textcolor{gray}{\textit{55.3}} &\textcolor{gray}{\textit{$\downarrow$ 2.87}} \\
      5 & \hspace{1.5mm} + Native resolution training & 56.7 & 59.5 & 67.7 & \textcolor{ForestGreen}{$\uparrow$ 3.33} \\
      6 & \hspace{1.5mm} + Increase resolution to 1024px in Stage-II & 56.2 & 59.9 & 68.8 & \textcolor{ForestGreen}{$\uparrow$ 0.33}\\
      7 & \hspace{1.5mm} + Scale up training data & 62.0 & 65.2 & 75.0  & \textcolor{ForestGreen}{$\uparrow$ 5.77}\\
      \rowcolor{RoyalBlue!3}
      8 & \hspace{1.5mm} + Alignment Loss & 61.9 & 66.7 & 75.9 & \textcolor{ForestGreen}{$\uparrow$ 0.77} \\
      \midrule
      9 & Replace with DINOv2-Large in \#7  & 58.9 & 61.5 & 
      68.5 & -\\
      \rowcolor{RoyalBlue!3}
      10 & \hspace{1.5mm} + Alignment Loss & 60.0 & 66.1 & 71.0 & \textcolor{ForestGreen}{$\uparrow$ 2.73} \\
      \bottomrule
      \end{tabular}
    \label{tab:ab_recipe}
\end{table}
\vspace{-3mm}

\subsection{Semantic segmentation}

\noindent\textbf{Setup.}
We evaluate the local-view semantic quality of \system by semantic segmentation on ADE20K~\cite{ade20k}, utilizing UperNet~\cite{upernet} by head tuning. Specifically, we freeze the backbone and only train the UperNet head at fixed 504px and 672px. To further analyze the segmentation performance of our model, we evaluate SigLIP 2 NaFlex variant~\cite{siglip2}, which also support native resolution inputs, and LLaVA-SigLIP which is from the checkpoint of LLaVA-OV-SI-0.5B model. All our vision models are from Stage-II of \system-MM-1B.

\noindent\textbf{Results.}
As shown in \cref{tab:main_seg}, for vision-only pre-training DINOv2, \system preserves the segmentation capability; for vision-language pre-training SigLIP, \system significantly enhances its pixel-level understanding ability; for AIMv2 which has visual reconstruction process, \system-AIMv2 brings the ability of resolution extrapolation, and outperforms AIMv2 on 672px by a large margin.
\vspace{-1mm}

\section{Ablation study}
\vspace{-1mm}

\label{sec:ablation}
We select AI2D~\cite{ai2d}, ChartQA~\cite{chartqa}, and DocVQA~\cite{docvqa} as the main evaluation sets for ablation studies, as these widely adopted, text-rich and fine-grained benchmarks provide a direct assessment of the vision model while minimizing interference from the prior knowledge of the language model~\cite{cambrain}. 

\noindent\textbf{Improved training recipe.}
\cref{tab:ab_recipe} presents the roadmap from the commonly used LLaVA-1.5-like architecture to our proposed \system. We begin our experiments with a data combination strategy basically following LLaVA-OV~\cite{llavaov}, while replacing the instruction-tuning data with 0.8M LLaVA-NeXT-SFT data~\cite{llavanext}. We adopt SigLIP-So400M~\cite{siglip} as the vision encoder and Qwen2.5-0.5B~\cite{qwen2.5} as the language decoder. Additionally, we progressively increase the input resolution across stages, using 384px, 576px, and 768px, respectively, by interpolation to the fixed-size position embeddings.

\begin{table}[t]
  \centering
  \caption{\textbf{Ablation on technique designs.}}
\vspace{-1mm}
  \begin{subtable}[t]{0.45\linewidth}
    \centering
      \tablestyle{0.8pt}{1}
      \caption{ \textbf{Ablation on \ourpe.}}
      \begin{tabular}{l  c  c  c  c}
      \toprule
      \# & Res. & AI2D & CQA & DVQA \\
      \midrule
      Learned PE & 384 & 48.8 &22.8 & 24.3  \\
      Learned PE & 768 & 47.2 &28.8 & 29.9  \\
      RoPE-2D & 768 & 47.5 &8.24 & 11.9  \\
      \rowcolor{RoyalBlue!3}
      \ourpe& 768 & 48.0 & 32.2 & 33.2 \\
      \midrule
       \multicolumn{5}{c}{\textit{Scale up training data to 8M}} \\ 
      RoPE-2D & Native & 57.8 & 29.5 & 34.3 \\
      \rowcolor{RoyalBlue!3}
      \ourpe& Native & 61.9 & 66.7 & 75.9 \\
      \bottomrule
      \end{tabular}
    \label{tab:ab_pe}
  \end{subtable}
  \hfill
  \begin{subtable}[t]{0.5\linewidth}
 \centering
      \tablestyle{1pt}{1.14}
      \caption{ \textbf{Ablation on Alignment Loss.}}
      \begin{tabular}{l  c  c  c}
      \toprule
      \# & AI2D & CQA & DVQA \\
      \midrule
      Contrastive Loss & 51.1 & 45.1 & 39.7 \\
      \midrule
      \rowcolor{RoyalBlue!3}
      Alignment Loss & 52.3 & 46.4 & 43.4 \\
      w/o $\mathbf{u}_W$ in \cref{eq:algloss} & 51.0 & 44.0 & 35.4 \\
      w/o word embedding &  50.8 & 44.3 & 35.9 \\
      w/o Sinkhorn-Knopp &  50.7 & 45.6 & 37.3 \\
      w/o Alignment Loss &  51.0 & 45.0 & 36.9 \\
      \bottomrule
      \end{tabular}
    \label{tab:ab_align}
  \end{subtable}
\end{table}

\noindent\textbf{Effectiveness of \ourpe.}
We utilize Qwen2-0.5B~\cite{qwen2} and SigLIP-So400M~\cite{siglip} to directly be fine-tuned on LLaVA-NeXT-SFT data~\cite{llavanext} for one epoch in three different settings: (1) Only learned position embedding with interpolation at pre-trained resolution 384px and higher resolution 768px, (2) only RoPE-2D and directly removing original position embedding~\cite{qwen2vl} at 768px, and (3) \ourpe at 768px. As shown in \cref{tab:ab_pe}, although the learned position embedding can obtain a certain capability to process high-resolution inputs through interpolation, \ourpe can further unleash the performance by a large margin. Moreover, the performance degradation is observed when directly using RoPE-2D like previous methods, \eg Qwen2VL~\cite{qwen2vl}. To further investigate the failure of RoPE-2D without learned positional embedding, we scale the training data to 8M and replace LLM with Qwen2.5-0.5B~\cite{qwen2.5}, yet the model still exhibits significant performance degradation, suggesting that it is neither a data-efficient nor a training-friendly approach compared to \ourpe.

\noindent\textbf{Design choices of Alignment Loss.}
We train Qwen2.5-0.5B~\cite{qwen2.5} and SigLIP-So400M~\cite{siglip} with LLaVA-NeXT~\cite{llavanext} dataset combination to evaluate different auxiliary losses and components in Alignment Loss. As shown in \cref{tab:ab_align}, contrastive loss typically demands large batch sizes for effective optimization (\eg, 65K in CoCa~\cite{Yu2022CoCaCC}, 164K in InternVL~\cite{internvl}, and for comparison, our mini-batch size is 32), resulting in substantial computational overhead. In addition, we investigate three variants of our Alignment Loss: (1) Without the prior distribution of word embeddings; (2) Replacing the word embedding prototypes with learnable prototypes following DINO~\cite{dino}; (3) Replacing the iterative Sinkhorn-Knopp algorithm with simple softmax function. The results indicate that the Sinkhorn-Knopp algorithm, when combined with word embedding prototypes and their prior distribution, effectively aligns visual features with the LLM space.

\section{Related work}
\label{sec:related}

\noindent\textbf{Vision foundation models.}
Large-scale vision pre-training has achieved remarkable progress~\cite{he2020moco,chen2020simclr,clip,siglip,siglip2,aimv2}, with Vision Transformers~\cite{vit} becoming the dominant backbone for visual understanding. Existing approaches mainly fall into two categories: vision-only pre-training and vision-language pre-training. Vision-only methods learn representations through contrastive learning~\cite{dino,dinov2,he2020moco,chen2020simclr} or masked image reconstruction~\cite{mae,beit,videomae}, while vision-language approaches~\cite{clip,siglip,siglip2,aimv2} align visual and textual representations using large-scale image-text pairs. Recent models such as AIMv2~\cite{aimv2} further adopt multimodal pre-training for vision encoders. However, existing VFMs still struggle with flexible input resolutions and effective cross-modal alignment in multimodal understanding scenarios.

\noindent\textbf{Multimodal pre-training.}
CLIP~\cite{clip} and its follow-ups~\cite{ALIGN2021ICML,zhang2024longclip,schuhmann2022laionb,spec} demonstrate the effectiveness of aligning vision and language representations into a shared semantic space for open-set classification and retrieval. However, such approaches often face challenges on fine-grained tasks, \eg segmentation and detailed captioning~\cite{clipself,lin2023sphinx}, due to their holistic visual representations and limited resolution handling. More recent methods, including Flamingo~\cite{alayrac2022flamingo}, CoCa~\cite{Yu2022CoCaCC}, BLIP~\cite{li2022blip}, BLIP-2~\cite{li2023blip}, and InstructBLIP~\cite{dai2024instructblip}, introduce cross-attention mechanisms for image-conditioned text generation. Meanwhile, LLaVA~\cite{llava,Liu2023ImprovedBW} and subsequent works~\cite{llavanext,deepstack,gao2024sphinx,lin2023vila,internvl,bai2023qwen,qwen2vl,wang2023see} leverage pretrained LLMs together with vision encoders to improve multimodal understanding through enhanced visual token processing, image tiling, and large-scale training.

In contrast to these approaches, we focus on enhancing the foundational visual capabilities of VFMs through a text-supervised generative pre-training paradigm. Furthermore, we improve flexible-resolution visual encoding by incorporating RoPE-2D and introduce an additional alignment objective to better bridge visual and language representations. As a result, our method achieves strong performance on both multimodal understanding and traditional vision tasks.
\vspace{-3mm}
\section{Conclusion}
We introduced \system, a continual multimodal pre-training pipeline to tackle fixed resolution and modality gap problems under LMM framework, for both vision-language pre-training and vision-only pre-training models. Specifically, we proposed \ourpe and Alignment Loss, which efficiently adapt to native resolution inputs with light continual training data and are better suited for the LLM text space. Through extensive experiments, we demonstrated that our approach achieved state-of-the-art performance on multimodal understanding benchmarks, and the performance of the VFMs in other downstream visual tasks is preserved

\section*{Acknowledgments}
This work is supported by the National Natural Science Foundation of China (Grant No. 62521004) and the Science and Technology Commission of Shanghai Municipality (No. 24511103100).

{
\tiny
\renewcommand{\baselinestretch}{0.92}\selectfont
\bibliographystyle{IEEEtran}
    \bibliography{main}

@String(CVPR= {IEEE Conf. Comput. Vis. Pattern Recog.})

@String(ICCV= {Int. Conf. Comput. Vis.})

@String(ECCV= {Eur. Conf. Comput. Vis.})

@String(ICLR = {Int. Conf. Learn. Represent.})

@String(CVPR  = {CVPR})

@String(ICCV  = {ICCV})

@String(ECCV  = {ECCV})

@String(ICLR  = {ICLR})

@inproceedings{llava,
  title={Visual instruction tuning},
  author={Liu, Haotian and Li, Chunyuan and Wu, Qingyang and Lee, Yong Jae},
  booktitle = {NeurIPS},
  year={2023}
}

@article{llavaov,
  title={LLaVA-OneVision: Easy Visual Task Transfer}, 
  author={Bo Li and Yuanhan Zhang and Dong Guo and Renrui Zhang and Feng Li and Hao Zhang and Kaichen Zhang and Peiyuan Zhang and Yanwei Li and Ziwei Liu and Chunyuan Li},
  year={2024},
  journal={TMLR},
}

@article{dinov2,
  title={DINOv2: Learning Robust Visual Features without Supervision},
  author={Oquab, Maxime and Darcet, Timoth{\'e}e and Moutakanni, Th{\'e}o and Vo, Huy and Szafraniec, Marc and Khalidov, Vasil and Fernandez, Pierre and Haziza, Daniel and Massa, Francisco and El-Nouby, Alaaeldin and others},
  journal={TMLR},
  year={2024}
}

@inproceedings{clip,
  title={Learning transferable visual models from natural language supervision},
  author={Radford, Alec and Kim, Jong Wook and Hallacy, Chris and Ramesh, Aditya and Goh, Gabriel and Agarwal, Sandhini and Sastry, Girish and Askell, Amanda and Mishkin, Pamela and Clark, Jack and others},
  booktitle={ICML},
  year={2021},
}

@inproceedings{siglip,
  title={Sigmoid loss for language image pre-training},
  author={Zhai, Xiaohua and Mustafa, Basil and Kolesnikov, Alexander and Beyer, Lucas},
  booktitle={ICCV},
  year={2023}
}

@inproceedings{densefusion,
  title={Densefusion-1m: Merging vision experts for comprehensive multimodal perception},
  author={Li, Xiaotong and Zhang, Fan and Diao, Haiwen and Wang, Yueze and Wang, Xinlong and Duan, Ling-Yu},
  year={2024},
  booktitle = {NeurIPS},
}

@article{aimv2,
      title={Multimodal Autoregressive Pre-training of Large Vision Encoders}, 
      author={Enrico Fini and Mustafa Shukor and Xiujun Li and Philipp Dufter and Michal Klein and David Haldimann and Sai Aitharaju and Victor Guilherme Turrisi da Costa and Louis Béthune and Zhe Gan and Alexander T Toshev and Marcin Eichner and Moin Nabi and Yinfei Yang and Joshua M. Susskind and Alaaeldin El-Nouby},
      year={2024},
      eprint={2411.14402},
      archivePrefix={arXiv},
      primaryClass={cs.CV},
      journal={arXiv preprint arXiv:2411.14402}, 
}

@article{siglip2,
      title={SigLIP 2: Multilingual Vision-Language Encoders with Improved Semantic Understanding, Localization, and Dense Features}, 
      author={Michael Tschannen and Alexey Gritsenko and Xiao Wang and Muhammad Ferjad Naeem and Ibrahim Alabdulmohsin and Nikhil Parthasarathy and Talfan Evans and Lucas Beyer and Ye Xia and Basil Mustafa and Olivier Hénaff and Jeremiah Harmsen and Andreas Steiner and Xiaohua Zhai},
      year={2025},
      eprint={2502.14786},
      archivePrefix={arXiv},
      primaryClass={cs.CV},
      journal={arXiv preprint arXiv:2502.14786}, 
}

@article{qwen2vl,
      title={Qwen2-VL: Enhancing Vision-Language Model's Perception of the World at Any Resolution}, 
      author={Peng Wang and Shuai Bai and Sinan Tan and Shijie Wang and Zhihao Fan and Jinze Bai and Keqin Chen and Xuejing Liu and Jialin Wang and Wenbin Ge and Yang Fan and Kai Dang and Mengfei Du and Xuancheng Ren and Rui Men and Dayiheng Liu and Chang Zhou and Jingren Zhou and Junyang Lin},
      year={2024},
      eprint={2409.12191},
      archivePrefix={arXiv},
      primaryClass={cs.CV},
      journal={arXiv preprint arXiv:2409.12191}, 
}

@article{qwen2.5vl,
      title={Qwen2.5-VL Technical Report}, 
      author={Shuai Bai and Keqin Chen and Xuejing Liu and Jialin Wang and Wenbin Ge and Sibo Song and Kai Dang and Peng Wang and Shijie Wang and Jun Tang and Humen Zhong and Yuanzhi Zhu and Mingkun Yang and Zhaohai Li and Jianqiang Wan and Pengfei Wang and Wei Ding and Zheren Fu and Yiheng Xu and Jiabo Ye and Xi Zhang and Tianbao Xie and Zesen Cheng and Hang Zhang and Zhibo Yang and Haiyang Xu and Junyang Lin},
      year={2025},
      eprint={2502.13923},
      archivePrefix={arXiv},
      primaryClass={cs.CV},
      journal={arXiv preprint arXiv:2502.13923}, 
}

@article{qwen2.5,
      title={Qwen2.5 Technical Report}, 
      author={An Yang and Baosong Yang and Beichen Zhang and Binyuan Hui and Bo Zheng and Bowen Yu and Chengyuan Li and Dayiheng Liu and Fei Huang and Haoran Wei and Huan Lin and Jian Yang and Jianhong Tu and Jianwei Zhang and Jianxin Yang and Jiaxi Yang and Jingren Zhou and Junyang Lin and Kai Dang and Keming Lu and Keqin Bao and Kexin Yang and Le Yu and Mei Li and Mingfeng Xue and Pei Zhang and Qin Zhu and Rui Men and Runji Lin and Tianhao Li and Tianyi Tang and Tingyu Xia and Xingzhang Ren and Xuancheng Ren and Yang Fan and Yang Su and Yichang Zhang and Yu Wan and Yuqiong Liu and Zeyu Cui and Zhenru Zhang and Zihan Qiu},
      year={2025},
      eprint={2412.15115},
      archivePrefix={arXiv},
      primaryClass={cs.CL},
      journal={arXiv preprint arXiv:2412.15115}, 
}

@article{internvl,
      title={Expanding Performance Boundaries of Open-Source Multimodal Models with Model, Data, and Test-Time Scaling}, 
      author={Zhe Chen and Weiyun Wang and Yue Cao and Yangzhou Liu and Zhangwei Gao and Erfei Cui and Jinguo Zhu and Shenglong Ye and Hao Tian and Zhaoyang Liu and Lixin Gu and Xuehui Wang and Qingyun Li and Yimin Ren and Zixuan Chen and Jiapeng Luo and Jiahao Wang and Tan Jiang and Bo Wang and Conghui He and Botian Shi and Xingcheng Zhang and Han Lv and Yi Wang and Wenqi Shao and Pei Chu and Zhongying Tu and Tong He and Zhiyong Wu and Huipeng Deng and Jiaye Ge and Kai Chen and Kaipeng Zhang and Limin Wang and Min Dou and Lewei Lu and Xizhou Zhu and Tong Lu and Dahua Lin and Yu Qiao and Jifeng Dai and Wenhai Wang},
      year={2025},
      eprint={2412.05271},
      archivePrefix={arXiv},
      primaryClass={cs.CV},
      journal={arXiv preprint arXiv:2412.05271}, 
}

@article{qwen2,
      title={Qwen2 Technical Report}, 
      author={An Yang and Baosong Yang and Binyuan Hui and Bo Zheng and Bowen Yu and Chang Zhou and Chengpeng Li and Chengyuan Li and Dayiheng Liu and Fei Huang and Guanting Dong and Haoran Wei and Huan Lin and Jialong Tang and Jialin Wang and Jian Yang and Jianhong Tu and Jianwei Zhang and Jianxin Ma and Jianxin Yang and Jin Xu and Jingren Zhou and Jinze Bai and Jinzheng He and Junyang Lin and Kai Dang and Keming Lu and Keqin Chen and Kexin Yang and Mei Li and Mingfeng Xue and Na Ni and Pei Zhang and Peng Wang and Ru Peng and Rui Men and Ruize Gao and Runji Lin and Shijie Wang and Shuai Bai and Sinan Tan and Tianhang Zhu and Tianhao Li and Tianyu Liu and Wenbin Ge and Xiaodong Deng and Xiaohuan Zhou and Xingzhang Ren and Xinyu Zhang and Xipin Wei and Xuancheng Ren and Xuejing Liu and Yang Fan and Yang Yao and Yichang Zhang and Yu Wan and Yunfei Chu and Yuqiong Liu and Zeyu Cui and Zhenru Zhang and Zhifang Guo and Zhihao Fan},
      year={2024},
      eprint={2407.10671},
      archivePrefix={arXiv},
      primaryClass={cs.CL},
      journal={arXiv preprint arXiv:2407.10671}, 
}

@misc{rope2d,
  title={Transformer upgrade path: 17. insights into multimodal positional encoding},
  url={https://spaces.ac.cn/archives/10040},
  author={Su, Jianlin},
  year={2024}
}

@misc{llavanext,
    title={LLaVA-NeXT: Improved reasoning, OCR, and world knowledge},
    url={https://llava-vl.github.io/blog/2024-01-30-llava-next/},
    author={Liu, Haotian and Li, Chunyuan and Li, Yuheng and Li, Bo and Zhang, Yuanhan and Shen, Sheng and Lee, Yong Jae},
    year={2024}
}

@inproceedings{deepstack,
  title={DeepStack: Deeply Stacking Visual Tokens is Surprisingly Simple and Effective for LMMs},
  author={Meng, Lingchen and Yang, Jianwei and Tian, Rui and Dai, Xiyang and Wu, Zuxuan and Gao, Jianfeng and Jiang, Yu-Gang},
  booktitle={NeurIPS},
  year={2024}
}

@article{rope,
  title={Roformer: Enhanced transformer with rotary position embedding},
  author={Su, Jianlin and Ahmed, Murtadha and Lu, Yu and Pan, Shengfeng and Bo, Wen and Liu, Yunfeng},
  journal={Neurocomputing},
  year={2024}
}

@inproceedings{mae,
  title={Masked autoencoders are scalable vision learners},
  author={He, Kaiming and Chen, Xinlei and Xie, Saining and Li, Yanghao and Doll{\'a}r, Piotr and Girshick, Ross},
  booktitle={CVPR},
  year={2022}
}

@inproceedings{navit,
  title={Patch n’pack: Navit, a vision transformer for any aspect ratio and resolution},
  author={Dehghani, Mostafa and Mustafa, Basil and Djolonga, Josip and Heek, Jonathan and Minderer, Matthias and Caron, Mathilde and Steiner, Andreas and Puigcerver, Joan and Geirhos, Robert and Alabdulmohsin, Ibrahim M and others},
  booktitle={NeurIPS},
  year={2023}
}

@inproceedings{flexivit,
  title={Flexivit: One model for all patch sizes},
  author={Beyer, Lucas and Izmailov, Pavel and Kolesnikov, Alexander and Caron, Mathilde and Kornblith, Simon and Zhai, Xiaohua and Minderer, Matthias and Tschannen, Michael and Alabdulmohsin, Ibrahim and Pavetic, Filip},
  booktitle={CVPR},
  year={2023}
}

@inproceedings{resformer,
    author    = {Tian, Rui and Wu, Zuxuan and Dai, Qi and Hu, Han and Qiao, Yu and Jiang, Yu-Gang},
    title     = {ResFormer: Scaling ViTs With Multi-Resolution Training},
    booktitle = {CVPR},
    year      = {2023},
}

@inproceedings{vlap,
  title={BRIDGING VISION AND LANGUAGE SPACES WITH ASSIGNMENT PREDICTION},
  author={Park, Jungin and Lee, Jiyoung and Sohn, Kwanghoon},
  booktitle={ICLR},
  year={2024}
}

@inproceedings{chartqa,
  title={ChartQA: A Benchmark for Question Answering about Charts with Visual and Logical Reasoning},
  author={Masry, Ahmed and Do, Xuan Long and Tan, Jia Qing and Joty, Shafiq and Hoque, Enamul},
  booktitle={ACL Findings},
  year={2022}
}

@inproceedings{docvqa,
  title={Docvqa: A dataset for vqa on document images},
  author={Mathew, Minesh and Karatzas, Dimosthenis and Jawahar, CV},
  booktitle={WACV},
  year={2021}
}

@article{instit,
      title={Inst-IT: Boosting Multimodal Instance Understanding via Explicit Visual Prompt Instruction Tuning}, 
      author={Wujian Peng and Lingchen Meng and Yitong Chen and Yiweng Xie and Yang Liu and Tao Gui and Hang Xu and Xipeng Qiu and Zuxuan Wu and Yu-Gang Jiang},
      year={2024},
      eprint={2412.03565},
      archivePrefix={arXiv},
      primaryClass={cs.CV},
      journal={arXiv preprint arXiv:2412.03565}, 
}

@inproceedings{vit,
  title={An Image is Worth 16x16 Words: Transformers for Image Recognition at Scale},
  author={Dosovitskiy, Alexey and Beyer, Lucas and Kolesnikov, Alexander and Weissenborn, Dirk and Zhai, Xiaohua and Unterthiner, Thomas and  Dehghani, Mostafa and Minderer, Matthias and Heigold, Georg and Gelly, Sylvain and Uszkoreit, Jakob and Houlsby, Neil},
  booktitle={ICLR},
  year={2021}
}

@inproceedings{mindthegap,
  title={Mind the gap: Understanding the modality gap in multi-modal contrastive representation learning},
  author={Liang, Victor Weixin and Zhang, Yuhui and Kwon, Yongchan and Yeung, Serena and Zou, James Y},
  booktitle={NeurIPS},
  year={2022}
}

@article{pixtral,
      title={Pixtral 12B}, 
      author={Pravesh Agrawal and Szymon Antoniak and Emma Bou Hanna and Baptiste Bout and Devendra Chaplot and Jessica Chudnovsky and Diogo Costa and Baudouin De Monicault and Saurabh Garg and Theophile Gervet and Soham Ghosh and Amélie Héliou and Paul Jacob and Albert Q. Jiang and Kartik Khandelwal and Timothée Lacroix and Guillaume Lample and Diego Las Casas and Thibaut Lavril and Teven Le Scao and Andy Lo and William Marshall and Louis Martin and Arthur Mensch and Pavankumar Muddireddy and Valera Nemychnikova and Marie Pellat and Patrick Von Platen and Nikhil Raghuraman and Baptiste Rozière and Alexandre Sablayrolles and Lucile Saulnier and Romain Sauvestre and Wendy Shang and Roman Soletskyi and Lawrence Stewart and Pierre Stock and Joachim Studnia and Sandeep Subramanian and Sagar Vaze and Thomas Wang and Sophia Yang},
      year={2024},
      eprint={2410.07073},
      archivePrefix={arXiv},
      primaryClass={cs.CV},
      journal={arXiv preprint arXiv:2410.07073}, 
}

@inproceedings{sk,
  title={Sinkhorn distances: Lightspeed computation of optimal transport},
  author={Cuturi, Marco},
  booktitle={NeurIPS},
  year={2013}
}

@inproceedings{cc3m,
  title={Conceptual captions: A cleaned, hypernymed, image alt-text dataset for automatic image captioning},
  author={Sharma, Piyush and Ding, Nan and Goodman, Sebastian and Soricut, Radu},
  booktitle={ACL},
  year={2018}
}

@inproceedings{ai2d,
title={A diagram is worth a dozen images},
author={Kembhavi, Aniruddha and Salvato, Mike and Kolve, Eric and Seo, Minjoon and Hajishirzi, Hannaneh and Farhadi, Ali},
booktitle={ECCV},
year={2016},
}

@inproceedings{mmmu,
    title={MMMU: A Massive Multi-discipline Multimodal Understanding and Reasoning Benchmark for Expert AGI},
    author={Xiang Yue and Yuansheng Ni and Kai Zhang and Tianyu Zheng and Ruoqi Liu and Ge Zhang and Samuel Stevens and Dongfu Jiang and Weiming Ren and Yuxuan Sun and Cong Wei and Botao Yu and Ruibin Yuan and Renliang Sun and Ming Yin and Boyuan Zheng and Zhenzhu Yang and Yibo Liu and Wenhao Huang and Huan Sun and Yu Su and Wenhu Chen},
    booktitle={CVPR},
    year={2024},
}

@inproceedings{mmbench,
  title={Mmbench: Is your multi-modal model an all-around player?},
  author={Liu, Yuan and Duan, Haodong and Zhang, Yuanhan and Li, Bo and Zhang, Songyang and Zhao, Wangbo and Yuan, Yike and Wang, Jiaqi and He, Conghui and Liu, Ziwei and others},
  booktitle={ECCV},
  year={2024},
}

@misc{realworldqa,
  title={Grok-1.5 Vision Preview},
  url={https://x.ai/blog/grok-1.5v},
  author={xAI Team},
  year={2024}
}

@InProceedings{ALIGN2021ICML,
  title = {{Scaling Up Visual and Vision-Language Representation Learning With Noisy Text Supervision}},
  author = {Jia, Chao and Yang, Yinfei and Xia, Ye and Chen, Yi-Ting and Parekh, Zarana and Pham, Hieu and Le, Quoc and Sung, Yun-Hsuan and Li, Zhen and Duerig, Tom},
  year = {2021},
  booktitle = {ICML},
}

@inproceedings{Li2023BLIP2BL,
  title={BLIP-2: Bootstrapping Language-Image Pre-training with Frozen Image Encoders and Large Language Models},
  author={Junnan Li and Dongxu Li and Silvio Savarese and Steven C. H. Hoi},
  booktitle={ICML},
  year={2023},
}

@inproceedings{Yu2022CoCaCC,
  title={CoCa: Contrastive Captioners are Image-Text Foundation Models},
  author={Jiahui Yu and Zirui Wang and Vijay Vasudevan and Legg Yeung and Mojtaba Seyedhosseini and Yonghui Wu},
  booktitle={Trans. Mach. Learn. Res.},
  year={2022},
}

@article{Liu2023ImprovedBW,
  title={Improved Baselines with Visual Instruction Tuning},
  author={Haotian Liu and Chunyuan Li and Yuheng Li and Yong Jae Lee},
  journal={ArXiv},
  year={2023},
  volume={abs/2310.03744},
}

@misc{liu2024llavanext,
    title={LLaVA-NeXT: Improved reasoning, OCR, and world knowledge},
    author={Liu, Haotian and Li, Chunyuan and Li, Yuheng and Li, Bo and Zhang, Yuanhan and Shen, Sheng and Lee, Yong Jae},
    year={2024}
}

@article{li2023seed,
  title={Seed-bench: Benchmarking multimodal llms with generative comprehension},
  author={Li, Bohao and Wang, Rui and Wang, Guangzhi and Ge, Yuying and Ge, Yixiao and Shan, Ying},
  journal={arXiv preprint arXiv:2307.16125},
  year={2023}
}

@article{Touvron2023LLaMAOA,
  title={Llama: Open and efficient foundation language models},
  author={Touvron, Hugo and Lavril, Thibaut and Izacard, Gautier and Martinet, Xavier and Lachaux, Marie-Anne and Lacroix, Timoth{\'e}e and Rozi{\`e}re, Baptiste and Goyal, Naman and Hambro, Eric and Azhar, Faisal and others},
  journal={arXiv preprint arXiv:2302.13971},
  year={2023}
}

@article{Brown2020LanguageMA,
  title={Language models are few-shot learners},
  author={Brown, Tom and Mann, Benjamin and Ryder, Nick and Subbiah, Melanie and Kaplan, Jared D and Dhariwal, Prafulla and Neelakantan, Arvind and Shyam, Pranav and Sastry, Girish and Askell, Amanda and others},
  journal={NeurIPS},
  year={2020}
}

@article{lin2023vila,
  title={Vila: On pre-training for visual language models},
  author={Lin, Ji and Yin, Hongxu and Ping, Wei and Lu, Yao and Molchanov, Pavlo and Tao, Andrew and Mao, Huizi and Kautz, Jan and Shoeybi, Mohammad and Han, Song},
  journal={arXiv preprint arXiv:2312.07533},
  year={2023}
}

@article{radford2018improving,
  title={Improving language understanding by generative pre-training},
  author={Radford, Alec and Narasimhan, Karthik and Salimans, Tim and Sutskever, Ilya and others},
  year={2018},
}

@article{radford2019language,
  title={Language models are unsupervised multitask learners},
  author={Radford, Alec and Wu, Jeffrey and Child, Rewon and Luan, David and Amodei, Dario and Sutskever, Ilya and others},
  journal={OpenAI blog},
  year={2019}
}

@inproceedings{schuhmann2022laionb,
  title={{LAION}-5B: An open large-scale dataset for training next generation image-text models},
  author={Christoph Schuhmann and
          Romain Beaumont and
          Richard Vencu and
          Cade W Gordon and
          Ross Wightman and
          Mehdi Cherti and
          Theo Coombes and
          Aarush Katta and
          Clayton Mullis and
          Mitchell Wortsman and
          Patrick Schramowski and
          Srivatsa R Kundurthy and
          Katherine Crowson and
          Ludwig Schmidt and
          Robert Kaczmarczyk and
          Jenia Jitsev},
  booktitle={NeurIPS},
  year={2022},
}

@article{alayrac2022flamingo,
  title={Flamingo: a visual language model for few-shot learning},
  author={Alayrac, Jean-Baptiste and Donahue, Jeff and Luc, Pauline and Miech, Antoine and Barr, Iain and Hasson, Yana and Lenc, Karel and Mensch, Arthur and Millican, Katherine and Reynolds, Malcolm and others},
  journal={NeurIPS},
  year={2022}
}

@inproceedings{li2022blip,
  title={Blip: Bootstrapping language-image pre-training for unified vision-language understanding and generation},
  author={Li, Junnan and Li, Dongxu and Xiong, Caiming and Hoi, Steven},
  booktitle={ICML},
  year={2022},
}

@inproceedings{li2023blip,
  title={Blip-2: Bootstrapping language-image pre-training with frozen image encoders and large language models},
  author={Li, Junnan and Li, Dongxu and Savarese, Silvio and Hoi, Steven},
  booktitle={ICML},
  year={2023},
}

@article{dai2024instructblip,
  title={Instructblip: Towards general-purpose vision-language models with instruction tuning},
  author={Dai, Wenliang and Li, Junnan and Li, Dongxu and Tiong, Anthony Meng Huat and Zhao, Junqi and Wang, Weisheng and Li, Boyang and Fung, Pascale N and Hoi, Steven},
  journal={NeurIPS},
  year={2024}
}

@article{lin2023sphinx,
  title={Sphinx: The joint mixing of weights, tasks, and visual embeddings for multi-modal large language models},
  author={Lin, Ziyi and Liu, Chris and Zhang, Renrui and Gao, Peng and Qiu, Longtian and Xiao, Han and Qiu, Han and Lin, Chen and Shao, Wenqi and Chen, Keqin and others},
  journal={arXiv preprint arXiv:2311.07575},
  year={2023}
}

@article{gao2024sphinx,
  title={SPHINX-X: Scaling Data and Parameters for a Family of Multi-modal Large Language Models},
  author={Gao, Peng and Zhang, Renrui and Liu, Chris and Qiu, Longtian and Huang, Siyuan and Lin, Weifeng and Zhao, Shitian and Geng, Shijie and Lin, Ziyi and Jin, Peng and others},
  journal={arXiv preprint arXiv:2402.05935},
  year={2024}
}

@article{zong2024mova,
  title={MoVA: Adapting Mixture of Vision Experts to Multimodal Context},
  author={Zong, Zhuofan and Ma, Bingqi and Shen, Dazhong and Song, Guanglu and Shao, Hao and Jiang, Dongzhi and Li, Hongsheng and Liu, Yu},
  journal={arXiv preprint arXiv:2404.13046},
  year={2024}
}

@inproceedings{mathew2022infographicvqa,
  title={Infographicvqa},
  author={Mathew, Minesh and Bagal, Viraj and Tito, Rub{\`e}n and Karatzas, Dimosthenis and Valveny, Ernest and Jawahar, CV},
  booktitle={WACV},
  year={2022}
}

@inproceedings{singh2019towards,
  title={Towards vqa models that can read},
  author={Singh, Amanpreet and Natarajan, Vivek and Shah, Meet and Jiang, Yu and Chen, Xinlei and Batra, Dhruv and Parikh, Devi and Rohrbach, Marcus},
  booktitle={CVPR},
  year={2019}
}

@article{bai2023qwen,
  title={Qwen-vl: A frontier large vision-language model with versatile abilities},
  author={Bai, Jinze and Bai, Shuai and Yang, Shusheng and Wang, Shijie and Tan, Sinan and Wang, Peng and Lin, Junyang and Zhou, Chang and Zhou, Jingren},
  journal={arXiv preprint arXiv:2308.12966},
  year={2023}
}

@inproceedings{hudson2019gqa,
  title={Gqa: A new dataset for real-world visual reasoning and compositional question answering},
  author={Hudson, Drew A and Manning, Christopher D},
  booktitle={CVPR},
  year={2019}
}

@inproceedings{schwenk2022okvqa,
  title={A-okvqa: A benchmark for visual question answering using world knowledge},
  author={Schwenk, Dustin and Khandelwal, Apoorv and Clark, Christopher and Marino, Kenneth and Mottaghi, Roozbeh},
  booktitle={ECCV},
  year={2022},
}

@article{wang2023see,
  title={To see is to believe: Prompting gpt-4v for better visual instruction tuning},
  author={Wang, Junke and Meng, Lingchen and Weng, Zejia and He, Bo and Wu, Zuxuan and Jiang, Yu-Gang},
  journal={arXiv preprint arXiv:2311.07574},
  year={2023}
}

@article{llama3,
  title = {The Llama 3 Herd of Models},
  author = {Dubey, Avijit and Grattafiori, Aaron and others},
  journal = {arXiv preprint arXiv:2407.21783},
  year = {2024},
  url = {https://arxiv.org/abs/2407.21783}
}

@article{zhang2024longclip,
  title = {{Long-CLIP: Unlocking the Long-Text Capability of CLIP}},
  author = {Zhang, Beichen and Zhang, Pan and Dong, Xiaoyi and Zang, Yuhang and Wang, Jiaqi},
  journal = {arXiv preprint arXiv:2403.15378},
  year = {2024},
  url = {https://arxiv.org/abs/2403.15378}
}

@inproceedings{he2020moco,
  title     = {Momentum Contrast for Unsupervised Visual Representation Learning},
  author    = {He, Kaiming and Fan, Haoqi and Wu, Yuxin and Xie, Saining and Girshick, Ross},
  booktitle = {CVPR},
  year      = {2020},
}

@inproceedings{chen2020simclr,
  title     = {A Simple Framework for Contrastive Learning of Visual Representations},
  author    = {Chen, Ting and Kornblith, Simon and Norouzi, Mohammad and Hinton, Geoffrey},
  booktitle = {ICML},
  year      = {2020},
}

@inproceedings{dino,
  title={Emerging properties in self-supervised vision transformers},
  author={Caron, Mathilde and Touvron, Hugo and Misra, Ishan and J{\'e}gou, Herv{\'e} and Mairal, Julien and Bojanowski, Piotr and Joulin, Armand},
  booktitle={CVPR},
  year={2021}
}

@inproceedings{beit,
  title={BEiT: BERT Pre-Training of Image Transformers},
  author={Bao, Hangbo and Dong, Li and Piao, Songhao and Wei, Furu},
  booktitle={ICLR},
year={2021}
}

@article{videomae,
  title={Videomae: Masked autoencoders are data-efficient learners for self-supervised video pre-training},
  author={Tong, Zhan and Song, Yibing and Wang, Jue and Wang, Limin},
  journal={NeurIPS},
  year={2022}
}

@inproceedings{clipself,
  title={CLIPSelf: Vision Transformer Distills Itself for Open-Vocabulary Dense Prediction},
  author={Wu, Size and Zhang, Wenwei and Xu, Lumin and Jin, Sheng and Li, Xiangtai and Liu, Wentao and Loy, Chen Change},
  booktitle={ICLR},
  year={2024}
}

@inproceedings{xiong2024effective,
  title={Effective Long-Context Scaling of Foundation Models},
  author={Xiong, Wenhan and Liu, Jingyu and Molybog, Igor and Zhang, Hejia and Bhargava, Prajjwal and Hou, Rui and Martin, Louis and Rungta, Rashi and Sankararaman, Karthik Abinav and Oguz, Barlas and others},
  booktitle={NAACL},
  year={2024}
}

@inproceedings{imagenet,
  title={Imagenet: A large-scale hierarchical image database},
  author={Deng, Jia and Dong, Wei and Socher, Richard and Li, Li-Jia and Li, Kai and Fei-Fei, Li},
  booktitle={CVPR},
  year={2009},
}

@inproceedings{ade20k,
    title={Scene Parsing through ADE20K Dataset},
    author={Zhou, Bolei and Zhao, Hang and Puig, Xavier and Fidler, Sanja and Barriuso, Adela and Torralba, Antonio},
    booktitle={CVPR},
    year={2017}
}

@inproceedings{upernet,
  title={Unified perceptual parsing for scene understanding},
  author={Xiao, Tete and Liu, Yingcheng and Zhou, Bolei and Jiang, Yuning and Sun, Jian},
  booktitle={ECCV},
  year={2018}
}

@article{mme,
  title={MME: A Comprehensive Evaluation Benchmark for Multimodal Large Language Models},
  author={Fu, Chaoyou and Chen, Peixian and Shen, Yunhang and Qin, Yulei and Zhang, Mengdan and Lin, Xu and Yang, Jinrui and Zheng, Xiawu and Li, Ke and Sun, Xing and others},
  journal={arXiv preprint arXiv:2306.13394},
  year={2023}
}

@article{deepseekvl,
      title={DeepSeek-VL: Towards Real-World Vision-Language Understanding}, 
      author={Haoyu Lu and Wen Liu and Bo Zhang and Bingxuan Wang and Kai Dong and Bo Liu and Jingxiang Sun and Tongzheng Ren and Zhuoshu Li and Hao Yang and Yaofeng Sun and Chengqi Deng and Hanwei Xu and Zhenda Xie and Chong Ruan},
      year={2024},
      journal={arXiv preprint arXiv:2403.05525},
      archivePrefix={arXiv},
      primaryClass={cs.AI},
      url={https://arxiv.org/abs/2403.05525}, 
}

@article{cambrain,
      title={Cambrian-1: A Fully Open, Vision-Centric Exploration of Multimodal LLMs}, 
      author={Shengbang Tong and Ellis Brown and Penghao Wu and Sanghyun Woo and Manoj Middepogu and Sai Charitha Akula and Jihan Yang and Shusheng Yang and Adithya Iyer and Xichen Pan and Ziteng Wang and Rob Fergus and Yann LeCun and Saining Xie},
      year={2024},
      eprint={2406.16860},
      archivePrefix={arXiv},
      primaryClass={cs.CV},
      journal={arXiv preprint arXiv:2406.16860}, 
}

@inproceedings{spec,
    author    = {Peng, Wujian and Xie, Sicheng and You, Zuyao and Lan, Shiyi and Wu, Zuxuan},
    title     = {Synthesize Diagnose and Optimize: Towards Fine-Grained Vision-Language Understanding},
    booktitle = {CVPR},
    year      = {2024},
}

@inproceedings{ropevision,
  title={Rotary position embedding for vision transformer},
  author={Heo, Byeongho and Park, Song and Han, Dongyoon and Yun, Sangdoo},
  booktitle={ECCV},
  year={2024},
}

@inproceedings{lcl,
  title={Vision model pre-training on interleaved image-text data via latent compression learning},
  author={Yang, Chenyu and Zhu, Xizhou and Zhu, Jinguo and Su, Weijie and Wang, Junjie and Dong, Xuan and Wang, Wenhai and Li, Bin and Zhou, Jie and Qiao, Yu and others},
  booktitle={NeurIPS},
  year={2024}
}
}

\clearpage
\appendix

\section{More results}
\label{sec:appendix_more_results}
\begin{table}[!ht]
      \centering
      \tablestyle{8pt}{1.2}
      \caption{\textbf{Ablation on the effectiveness of Alignment loss.} We utilize Qwen2-0.5B~\cite{qwen2} and DINOv2-Large~\cite{dinov2}, and employ LLaVA-NeXT-SFT~\cite{llavanext} as training data of Stage-III.}
      \vspace{0.1in}
      \begin{tabular}{ l  c  c  c c }
      \toprule
       Recipe & AI2D & ChartQA & DocVQA  \\
      \midrule
      No Alignment Loss & 52.8 & 55.6 & 60.9 \\ 
      Only fixed res. & 54.0 & 59.6 & 68.4 \\ 
      Both fixed \& native res. & 54.1 & 60.2 & 67.5 \\ 
      \bottomrule
      \end{tabular}
    \label{tab:ab_e_align}
\end{table}

\noindent\textbf{Effectiveness of Alignment Loss.}
To further evaluate the effectiveness of Alignment Loss, we utilize Qwen2.5-0.5B~\cite{qwen2} and DINOv2-Large~\cite{dinov2}, and replace the training data of Stage-III with LLaVA-NeXT-SFT data~\cite{llavanext} for rapid evaluation in three different settings: (1) without Alignment Loss, (2) with Alignment Loss when the inputs are fixed resolution in Stage-I, II and (3) with Alignment Loss during entire Stage-I, II, including native resolution training. As shown in \cref{tab:ab_e_align}, Alignment Loss is particularly beneficial for text-rich tasks, and generally, the longer it is applied, the better the performance. Therefore, we use Alignment Loss in both entire Stage I and Stage II by default.

\begin{table}[!h]
      \centering
      \tablestyle{6pt}{1.2}
      \caption{\textbf{Ablation on vision foundation models.} We ablate the vision foundation models of different model sizes and pretraining tasks.}
      \vspace{0.1in}
      \begin{tabular}{ l  c  c  c c}
      \toprule
      Model  & AI2D & ChartQA & DocVQA & MMMU\\
      \midrule
      SigLIP-Base~\cite{siglip}  & 60.2 & 61.6 & 71.0 & 33.6\\
      SigLIP-So400M~\cite{siglip} & 61.9 & 66.7 & 75.9 & 33.0\\
      DINOv2-Large~\cite{dinov2} & 60.0 & 66.1 & 71.0 & 30.4 \\
      CLIP-Large~\cite{clip} & 61.6 & 67.3 & 78.6 & 33.1 \\
      AIMv2-Large~\cite{aimv2} & 64.9 & 68.3 & 75.8 & 35.7 \\
      \bottomrule
      \end{tabular}
    \label{tab:ab_ve}
\end{table}

\noindent\textbf{Comparisons with different VFMs.}
As shown in \cref{tab:ab_ve}, we explore different VFMs with SigLIP-Base, SigLIP-So400M, DINOv2-Large, CLIP-Large-336px and AIMv2-Large-336px. The results demonstrate that our method is applicable not only to different pre-training objectives but also to various model sizes.

\begin{figure}[h!]
    \centering
    \includegraphics[width=0.65\linewidth]{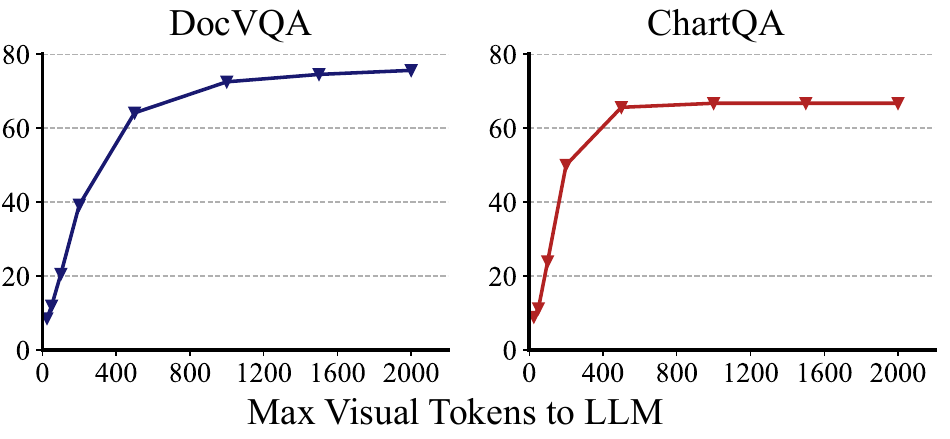}
    \caption{\textbf{Varying the image resolution during inference.} We investigate the impact of image resolution on DocVQA~\cite{docvqa} and ChartQA~\cite{chartqa} by \system-MM.} 
    \label{fig:plot}
\end{figure}
\noindent\textbf{Varying the resolution during inference.}
To investigate the impact of image resolution on resolution-sensitive tasks, we observe the changes in scores on DocVQA~\cite{docvqa} and ChartQA~\cite{chartqa} by limiting the max number of visual tokens to LLM. As shown in \cref{fig:plot}, as the input resolution increases, the performance of both tasks gradually improves, which demonstrates the effectiveness of our approach for varying resolutions, particularly the high resolution.

\begin{table}[!ht]
      \centering
      \tablestyle{6pt}{1.2}
      \caption{\textbf{Alignment Loss for visual downstream task.} We train SigLIP with and without Alignment Loss and test on classification and segmentation.}
      \vspace{0.1in}
      \begin{tabular}{ l  c  c  c c}
      \toprule
       & \multicolumn{2}{c}{Classification} & \multicolumn{2}{c}{Segmentation} \\
        & 224px &  448px &   504px &  672px\\
      \midrule
      \system-SigLIP  & 86.5 & 87.4 & 49.5 & 49.1 \\
      w/o Alignment Loss & 86.0 & 87.0 & 49.2 & 48.9\\
      \bottomrule
      \end{tabular}
    \label{tab:ab_align_on_vision}
\end{table}

\noindent\textbf{Alignment Loss for visual downstream task.}
As shown in \cref{tab:ab_ve}, we explore the effectiveness of Alignment Loss on visual downstream task like classification and segmentation. As shown in \cref{tab:ab_align_on_vision}, it consistently improves performance across tasks.

\begin{table}[!t]
      \centering
      \tablestyle{2pt}{1.2}
      \caption{ \textbf{Ablation on different language decoders.} We utilize Qwen2-0.5B~\cite{qwen2} and Qwen2.5-0.5B as different language decoders. L.D. denotes language decoder.}
      \vspace{0.1in}
      
      \begin{tabular}{l  c    c  c}
      \toprule
      Method & L.D. &  ChartQA & DocVQA \\
      \midrule
      LLaVA-OV (SI)~\cite{llavaov} & Qwen2~\cite{qwen2} & 61.0 & 75.0  \\
      LLaVA-OV~\cite{llavaov} & Qwen2~\cite{qwen2} & 61.4 & 73.7  \\
      \system-MM & Qwen2~\cite{qwen2} & 63.9 & \textbf{76.7}  \\
      \system-MM & Qwen2.5~\cite{qwen2.5} & \textbf{65.2} & 75.0  \\
      \bottomrule
      \end{tabular}
        \label{tab:apen_lm}
    
\end{table}

\noindent\textbf{Different Language Decoders.}
We use different LLMs, Qwen2-0.5B~\cite{qwen2} and Qwen2.5-0.5B~\cite{qwen2.5}, as the language decoder to verify the generalizability of our method. As shown in \cref{tab:apen_lm}, on text-rich tasks, Qwen2 and Qwen2.5 show no significant differences. Moreover, compared to LLaVA-OV~\cite{llavaov}, \system-MM-Qwen2 still demonstrates strong performance. Notably, Alignment Loss is not utilized in these experiments.

\begin{table}[!ht]
      \centering
      \tablestyle{4pt}{0.95}
      \caption{ \textbf{Full set results of \cref{tab:ab_recipe}.} Our final recipe is highlighted in \colorbox{RoyalBlue!10}{blue}. Experiments are run using SigLIP-So400M with Qwen2.5-0.5B.}
      \vspace{0.1in}
      \begin{tabular}{l  l c  c  c  c ccccc}
      \toprule
      \# & & AI2D & ChartQA &  DocVQA & Inst-IT & GQA & MMMU & MMBench & RealWorldQA \\
      \midrule
      1 & & 50.2 & 29.9 & 24.1 & 41.3 & 56.6 & 31.6&46.5& 49.2 \\
      \rowcolor{gray!2}
      \textcolor{gray}
      2 &  & \textcolor{gray}{\textit{51.5}} & \textcolor{gray}{\textit{11.0}} & \textcolor{gray}{\textit{11.4}} & \textcolor{gray}{\textit{37.0}} & \textcolor{gray}{\textit{39.0}} & \textcolor{gray}{\textit{31.4}}& \textcolor{gray}{\textit{19.8}} & \textcolor{gray}{\textit{42.0}}\\
      3 &  & 55.4 & 56.9 & 61.6 & 44.7 & 60.3 & 32.0 & 53.2 & 53.9 \\
      \rowcolor{gray!2}
      \textcolor{gray}
      4 &  & \textcolor{gray}{\textit{54.6}} & \textcolor{gray}{\textit{55.4}} & \textcolor{gray}{\textit{55.3}} & \textcolor{gray}{\textit{43.5}} & \textcolor{gray}{\textit{59.6}} &\textcolor{gray}{\textit{32.9}} &\textcolor{gray}{\textit{51.9}} & \textcolor{gray}{\textit{53.2}} \\

      5 &  & 56.7 & 59.5 & 67.7 & 44.2 & 60.2 & 34.3 & 51.2 & 57.8 \\
      7 &  & 62.0 & 65.2 & 75.0 & 49.8 & 60.4 & 32.6 & 55.8 & 56.6 \\
      \rowcolor{RoyalBlue!3}
      8 &  & 61.9 & 66.7 & 75.9 & 50.1 & 60.4 & 33.0 & 56.4 & 58.3 \\
      \bottomrule
      \end{tabular}
    \label{tab:ab_recipe_full}
\end{table}
\noindent\textbf{Full set evaluation.} We report the full set results of \cref{tab:ab_recipe} in \cref{tab:ab_recipe_full}. For clarity and consistency, we retain the original numbering. 

\section{Limitations and Future Work}
\label{sec:limitation}
In this work, we do not systematically investigate the scaling laws of models and data. Our main experiments are conducted on 1B and 7B models, with the largest vision foundation model (VFM) being only 400M parameters and the training set limited to 8M samples. While the results are promising, it remains an open question whether the observed trends hold at scale. We leave these as promising directions to
explore in the future.

\section{Hyperparameters}
\label{sec:hyper}
\noindent\textbf{\system-MM.}
We outline the optimization hyperparameters used during \system-MM continual pre-training in \cref{tab:apen_mm}, and the hyperparameters not listed remain consistent with LLaVA-OneVision~\cite{llavaov}.
\begin{table}[!ht]
      \centering
      \tablestyle{10pt}{1.2}
      \caption{Detailed configuration for each training stage of our \system-MM models.}
      \vspace{0.1in}
      \begin{tabular}{l c c c c }
      \toprule
      \multirow{2}{*}{} & \multirow{2}{*}{Stage-I} & \multicolumn{2}{c}{Stage-II} & \multirow{2}{*}{Stage-III} \\
      & &Fixed & Native & \\
      \midrule
      Trainable & Adapter & Full Model & Full Model & Full Model\\
      Batch Size & 32$\times$8 & 32$\times$8 & 32$\times$8 & 16$\times$8\\
      LR$_{Adapter}$ & $1\times10^{-3}$ &$5\times10^{-3}$ & $5\times10^{-3}$ & $1\times10^{-5}$\\
      LR$_{VFM}$ & - &$1\times10^{-4}$ &$1\times10^{-4}$ &$2\times10^{-5}$ \\
      LR$_{LLM}$ & - &$2\times10^{-5}$ &$2\times10^{-5}$ &$1\times10^{-5}$ \\
      Epoch & 1 & 1 & 1 & 1 \\

      \bottomrule
      \end{tabular}
    \label{tab:apen_mm}
\end{table}

Besides, $\mathbf{u}_W(k)$ in \cref{eq:algloss} is $\frac{N_k}{\sum_i N_i}$, where $N_k$ is the number of the $k$-th word in Stage-II dataset, $\alpha$ in \cref{eq:allloss} is set to 0.05, and all temperature coefficients are 0.005.

\vspace{0.2in}
\noindent\textbf{Multimodal Understanding.}
The hyperaparmeters used for the instruction tuning are detailed in \cref{tab:apen_mu}. We tune \system-SigLIP and \system-DINOv2 with LLama 3.0 8B~\cite{llama3} on LLaVA SFT data~\cite{liu2024llavanext} for one epoch. In addition, we used a $2\times2$ downsampling adapter to unleash the high-resolution perception capability of \system-SigLIP and \system-DINOv2, while keeping the number of tokenens to LLM at 576 for a fair comparison.

\begin{table}[!ht]
    \centering
    \tablestyle{8pt}{1.2}
        \caption{Detailed configuration of \system-SigLIP, \system-AIMv2 and \system-DINOv2 in instruction tuning for multimodal understanding.}
      \vspace{0.2in}
    \begin{tabular}{l c}
        \toprule
         Training Config \\
         \midrule
         Optimizer & AdamW  \\
         Decoder peak learning rate & $1\times10^{-5}$ \\
         Decoder peak learning rate & $1\times10^{-5}$ \\
         Adapter peak learning rate & $8\times10^{-5}$ \\
         Minimum learning rate & 0 \\
         Learning rate schedule & cosine decay \\
         Batch size & 128 \\
         Iterations & 5197 \\
         Warmup ratio & 0.05 \\
         Transformations & PadToSquare, Resize \\
         \bottomrule
    \end{tabular}
    \label{tab:apen_mu}
\end{table}

\vspace{0.2in}
\noindent\textbf{Image Recognition.}
The hyperaparmeters used for frozen trunk classification on ImageNet-1k~\cite{imagenet} are detailed in \cref{tab:apen_ir}. The $mean$ and $std$ in Normalization of \system-SigLIP and \system-DINOv2 are [(0.5, 0.5, 0.5), (0.5, 0.5, 0.5)] and [(0.485, 0.456, 0.406), (0.229, 0.224, 0.225)] respectively, following the original SigLIP~\cite{siglip} and DINOv2~\cite{dinov2}. We use the same hyperaparmeters for all models and baselines. For AIMv2~\cite{aimv2}, the $mean$ and $std$ in Normalization are [(0.481, 0.458, 0.408), (0.269, 0.261, 0.276)], and the learning rate is $8\times10^{-5}$ for 224px and $2\times10^{-5}$ for 448px.
\begin{table}[!ht]
    \centering
    \tablestyle{5pt}{1}
    \caption{Detailed configuration of \system-SigLIP and \system-DINOv2 for classification.}
    \vspace{0.1in}
    \begin{tabular}{l c c}
        \toprule
         Training Config  & 224px & 448px \\
         \hline
         Optimizer & \multicolumn{2}{c}{AdamW}  \\
         Peak learning rate & $1\times10^{-4}$ & $1\times10^{-5}$\\
         Minimum learning rate & $2\times10^{-5}$ & $5\times10^{-6}$ \\
         Learning rate schedule & \multicolumn{2}{c}{cosine decay} \\
         Batch size & 1024 & 256 \\
         Weight decay & \multicolumn{2}{c}{0.05} \\
         Epochs & \multicolumn{2}{c}{10}\\
         Warmup epochs & \multicolumn{2}{c}{1} \\
         Augmentations: \\
         \hspace{0.05in} RandomResizedCrop \\
         \hspace{0.15in} size & 224px & 448px\\
         \hspace{0.15in} scale &\multicolumn{2}{c}{(0.08, 1.0)}\\
         \hspace{0.15in} ratio & \multicolumn{2}{c}{(0.75, 1.33)}\\
         \hspace{0.15in} interpolation & \multicolumn{2}{c}{Bicubic} \\
         \hspace{0.05in} RandomHorizontalFlip & \multicolumn{2}{c}{$p=0.5$}\\
         \hspace{0.05in} ToTensor \\
         \hspace{0.05in} Normalize & \multicolumn{2}{c}{follows SigLIP or DINOv2}\\
         \bottomrule
    \end{tabular}
    \label{tab:apen_ir}
\end{table}

\begin{table}[!ht]
    \centering
    \tablestyle{5pt}{1}
        \caption{Detailed configuration of \system-SigLIP and \system-DINOv2 for semantic segmentation.}
        \vspace{0.1in}
    \begin{tabular}{l c c}
        \toprule
         Training Config  & 504px & 672px \\
         \hline
         Optimizer & \multicolumn{2}{c}{AdamW}  \\
         Weight decay & \multicolumn{2}{c}{0.05} \\
         Peak learning rate & \multicolumn{2}{c}{$4\times10^{-5}$} \\
         Minimum learning rate & \multicolumn{2}{c}{0} \\
         Learning rate schedule & \multicolumn{2}{c}{poly decay} \\
         Batch size & \multicolumn{2}{c}{16} \\
         Iterations & \multicolumn{2}{c}{80K}\\
         Warmup iters & \multicolumn{2}{c}{1500} \\
         Augmentations: \\
         \hspace{0.05in} RandomResizedCrop & 504px & 672px \\
         \hspace{0.05in} RandomFlip & \multicolumn{2}{c}{$p=0.5$}\\
         \hspace{0.05in} PhotoMetricDistortion \\
         \hspace{0.05in} Normalize & \multicolumn{2}{c}{follows SigLIP or DINOv2}\\
         \bottomrule
    \end{tabular}
    \label{tab:apen_seg}
\end{table}

\vspace{0.2in}
\noindent\textbf{Semantic Segmentation.}
The hyperaparmeters used for semantic segmentation on ADE20K~\cite{ade20k} are detailed in \cref{tab:apen_seg}. The $mean$ and $std$ in Normalization of \system-SigLIP and \system-DINOv2 follow the original SigLIP~\cite{siglip} and DINOv2~\cite{dinov2}. The $mean$ and $std$ of AIMv2~\cite{aimv2} are [(0.481, 0.458, 0.408), (0.269, 0.261, 0.276)].
\end{document}